# Student Modeling using Case-Based Reasoning in Conventional Learning System

*Indriana Hidayah[1], Alvi Syahrina[2], Adhistya Erna Permanasari[3]*
Department of Electrical Engineering and Information Technology
Universitas Gadjah Mada
Yogyakarta, Indonesia
[1]indriana.hidayah@ugm.ac.id
[2]alvi.syahrina@te.gadjahmada.edu
[3]adhistya@ugm.ac.id

*Abstract*— Conventional face-to-face classrooms are still the main learning system applied in Indonesia. In assisting such conventional learning towards an optimal learning, formative evaluations are needed to monitor the progress of the class. This task can be very hard when the size of the class is large. Hence, this research attempted to create a classroom monitoring system based on student's data of Department of Electrical Engineering and Information Technology UGM. In order to achieve the goal, a student modeling using Case-Based Reasoning (CBR) was proposed. A generic student model based on jCOLIBRI 2.3 framework was developed. The model represented student's knowledge of a subject. The result showed that the system was able to store and retrieve student's data for suggestion of the current situation and formative evaluation for one of the subject in the Department.

## I. INTRODUCTION

Every classroom learning process has its own challenges. These challenges are often different depending on many factors. One possible cause is the composition of the attendants which has different backgrounds. Furthermore, having too many students in the class often make the teaching, learning, and evaluation process very hard. Yet, classroom monitoring is indeed very important to be done to ensure that students will successfully pass a course, hence, a monitoring system is required.

A model of effective instruction was written by Slavin (1995). The model is called QAIT model that consists of Quality, Appropriateness, Incentive, and Time [1]. Quality refers to the quality of instruction. Incentive refers to the degree to which the teacher makes sure that the students are well-motivated to work on the task. Time refers to the degree to which students are given the right amount of time to learn the material [1]. However the model's component that became the emphasis of this paper is Appropriateness which refers to the appropriateness level of instruction, which is the degree to which the teacher makes sure that the instruction is appropriate to the student's level of understanding. QAIT model suggests that a personal approach is needed to achieve an optimal learning result.

On the other hand, formative evaluation [2] is the evaluation process of an educational program while it is still in development, with the purpose of continually improving the program. Thus, implementing formative evaluation will optimize the learning result. However, the implementation will not be easy when the size of the class is large, i.e. consists of more than 70 students. To help the implementation of formative evaluation in a big classroom, Information technology can be used as a tool.

The determination of delivering education material that follows Slavin's Appropriateness criteria and ability to execute formative evaluation faces challenges when the size of the class is large. This paper proposed a framework on this problem by developing a student model using case-based reasoning technique. Based on individual student model, a classroom monitoring system can be performed and personalized recommendations were given to students as well as teachers to refine the learning process.

The rest of the paper is organized in the following way. The concept of student modeling is described in section II. Section III presents case-based reasoning method in general and jCOLIBRI as a framework based on CBR. Section IV describes the proposed framework. Result of the experiments and the analysis is presented in section V. Finally, Section VI concludes with a summary and a future plan.

## II. STUDENT MODELING

Student Modeling (SM) is defined as the process of acquiring knowledge about the student in order to provide services, adaptive content and personalized instructional flow/s according to specific student's requirements [3]. Even though, student modeling techniques have been applied in many eLearning systems, the techniques are rarely used in conventional classrooms. Most SMs are built to support classroom learning that utilizes web-based learning or Intelligent Tutoring System (ITS). However fewer SMs are built to support face-to-face classroom learning.

Various techniques have been used to represent student models such as rules, fuzzy logic, Bayesian networks (BN), and case-based reasoning (CBR). Bayesian network is among the most used techniques, thus, there are many resourceful researches. Previous work by Gonzalez, Burguillo and Llamas describes a qualitative comparison between SM using BN and CBR[4]. Generally BN is described as a complex technique that needs high computation and has a complex process in extracting knowledge, meanwhile CBR is said to have more advantages as it is easier to handle, renew and maintained. CBR based SM is proved to provide more evidence and reason when a student misconception happened. It also facilitates supervision of student by enabling the tutor to have a continuous view of student performance, including quantitative and qualitative information.

The result of this previous research became the foundation of choosing Case-Based Reasoning as a method to build a student model in this research.

### III. CASE-BASED REASONING

Case-Based Reasoning is a method to solve problem using solutions taken to solve previous problems [5]. This step is executed with the belief that the same problem will have the same solution. Rather than depending on the general knowledge of the problem or the relationship of the problem and the solution, CBR focuses more on using specific knowledge about the problem, situation and case that has been experienced.

CBR is a branch of artificial intelligence, where it is specifically related with automating reasoning using previous cases, problem definition for the current situation, and search of the previous problem and adapting the previous solution for a new problem. CBR is considered new to the field of problem-solving and machine learning.

CBR consists of four stages or known as 4R stage: Retrieve, Reuse, Revise and Retain. The cycle of the four stages is illustrated in Figure 1.

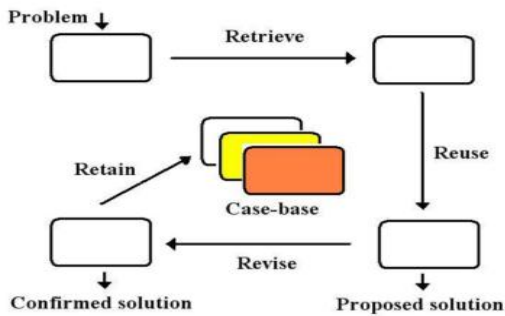

Figure 1. CBR Cycle [5]

The first stage is Retrieve. In this stage there is a process of extracting cases from one or a group of cases that is nearest to the new case. In computing the nearest case, the NN algorithm is used. The formula for NN is shown in (1).

$$similarity(T,S) = \frac{\sum_{i=1}^{n} f(T_i, S_i) \times w_i}{w_i} \quad (1)$$

where $T$ is the new case, $S$ is the case in the case base, $n$ is the number of attribute in every case, $i$ is the individual case between 1 to $n$, $f$ is the similarity function between case $T$ and case $S$, and $w$ is the weight assigned for $i$-th attribute that has value between $0 \leq w \leq 1$.

The second stage is Reuse where the solution of the previous case is reused to suggest solution to the new case. The third stage is Revise. In this stage, before storing the solution of the new case, the attributes' value of the case can be revised. Finally the last stage is Retain, where all of the information of the new case is stored in the case base.

*jCOLIBRI*

jCOLIBRI is an object oriented framework that is built to facilitate the design and implementation of a CBR system [6]. jCOLIBRI used Java programming language as a basis and JavaBeans for case representation. This framework is developed by an artificial intelligence group, GAIA, of University of Compultense, Madrid.

jCOLIBRI's main architecture consists of elements as follows:

1. Organization into three layers, persistence, core and presentation layers. Persistence is managed by connectors that access the persistence media and load the cases into different in-memory organizations. Core contains basic classes that has been previously defined.

2. Organization of the applications into precycle, cycle and postcyle. Moreover, new stages can be defined to be executed at different execution points. This add-on enables the development of maintenance or evaluation procedures.

3. Case structure consists of description, result, solution and justification.

### IV. A PROPOSED FRAMEWORK

The outcome of this research is a student model that is to be mapped into a system of case-based reasoning using jCOLIBRI as a tool. This system becomes the entity representation to evaluate case based reasoning of how effective it is to be a system for student modeling.

In this research a real classroom is selected where information such as student's data and course structure are learned. This information become the main materials to create a student model.

*A. Student Model*

After gathering information from a class, to represent the student's knowledge, a student model is created. This model is divided into components such as Student ID, cumulative GPA when the course is taken, grade of prerequisite courses, skills and/or experiences, competence that should be reached, exam or quiz result, and final grade. With Retrieval method using NN, the most similar previous case is obtained. Figure 2 shows the main tasks performed in the case-based reasoning technique to create the student model.

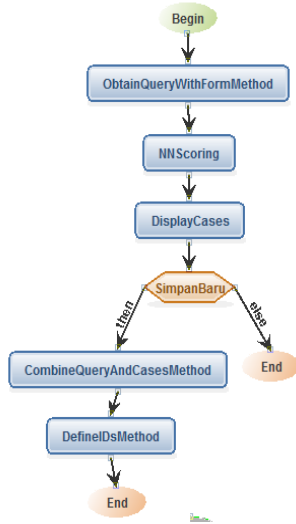

Figure 2. The tasks in CBR method in creating SM

Figure 3 illustrates the generic student model, where all of the components are stored in a case base of CBR system.

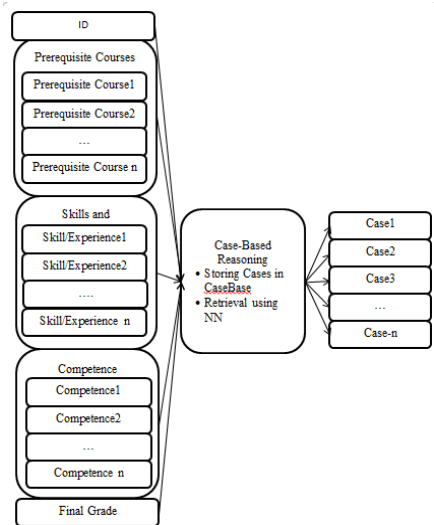

Figure 3. Student Model Representation

Student model that is specifically built for the Microprocessor System course has the component such as ID, GPA, grade of prerequisite course such as Digital Systems and Basic Programming, experiences and skill using assembly language and programming language, and designing an instrument. There are also quiz results, mid exam result and final grade of the course. These components became attributes of the cases that are organized into description, solution, result and justification.

After conducting this research, it is found that to get student's information from an "offline" class is harder than "online" class. The elements that are found in the offline class are only those that have numeric value in it, such as quiz or exam result in contrast to online class that can measure history or student's activity. In offline classes it is difficult to examine other things such as learning style.

*B. 4R Stages*

The 4R Stages of this system starts with the Retrieval stage where query of the new case is entered to find the similar cases. Figure 3a illustrates the query entry on the system. Users can enter value at the attribute field. From this query, NN scoring is used to find the most similar cases. These cases are shown in the Figure 3b.

Users can browse through these five most similar cases before choosing one from the five cases. By browsing through the cases, user can get some insight from the similar cases by observing their pattern. From the five cases users can choose one case where the attribute of the chosen case is reused, therefore the Reuse stage is conducted.

The third stage is the Revision stage. Here the components of the case are reviewed and the attributes are enabled to editing as shown in Figure 4a. If the new case has some differences to the selected case or some of the values needs to be updated, adjustments can be made in this stage.

Revision is also used to add more data into the case base if there is no case in the case base that is exactly the same with the new case. To add a new case into the case base a new ID needs to be defined.

The last stage is Retain. Retain is the activity where the new case is stored into the case base for future use as shown in Figure 4b. When the ID of the case is defined, either the same or new ID, the next button here is where the Retain stage is executed.

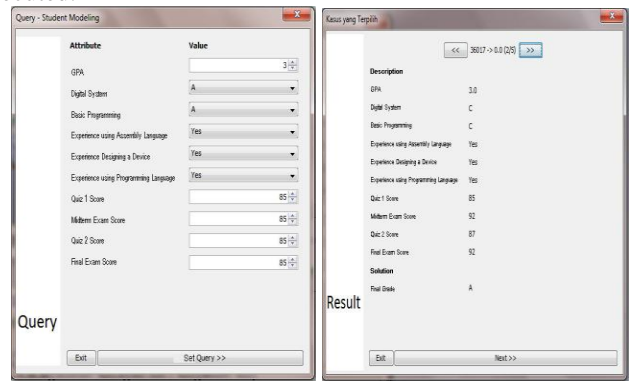

(a)                                          (b)

Figure 4. Query Panel and Result Panel

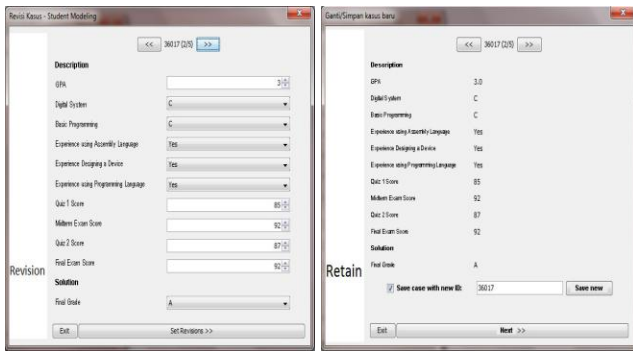

(a)            (b)

Figure 5. Revision Panel

*C. System Impact*

To examine the impact of the system as a tool to support formative evaluation, the following test is conducted.

There is a set of information about a student but only up to his mid-exam score ('UTS'). This student has a background with a good grade in his prerequisite courses but has low score on quiz and mid-exam.

After entering this data into the query, previous cases are obtained. There are four previous cases with the final grade of B and one previous case with A. It can be implied that there is a chance for this student to get good final grade. The case that has final grade of A shows significant change to the next quiz result and final exam. It means the student must do well in the next quiz and final exam. However if there is no significant change, like the other four results, most likely this student will get B.

## V. ADVANTAGES AND DISADVANTAGES

There are many advantages of CBR that is found in this research. Firstly is its simple computation. CBR's main computation is in its Retrieval stage on searching for the similar case. The rest activities in CBR only include storing and presenting data. Secondly, CBR do not look at any relation between the attributes. Some other SM techniques has relation between attributes and adds complexity to the system. Then CBR enables revision, which made the case base of the system stays updated.

However, some disadvantages are also found through this research. Firstly, the accuracy of the data depends on the case base. It means that all the data entered must be valid and the case base must stay updated. Secondly, one system can only be used for one course. The student model is general, but the system is specific to only one course. Other courses might have different attributes due to different teaching or different prerequisites.

jCOLIBRI as a framework also has many advantages and disadvantages that is found at the process of this research. The advantages of jCOLIBRI is that it uses Java, a language that is already common and have many IDE for programming. jCOLIBRI also have many previously written methods. However this can be the one of the drawbacks where system development depends on the availability of the method in the framework. For example, the data of the whole class cannot be observed because jCOLIBRI does not support to do so.

jCOLIBRI lacks in documentation. As a newly developed framework this is understandable. But there is no user forum, so there is no keeping track of those who use this framework for different purposes. This is seen as a major drawback in the programming field.

## VI. CONCLUSION

In this research project, a student model has been made based on CBR by using jCOLIBRI framework. Several conclusions can be drawn as the following.

- jCOLIBRI can facilitate 4R of CBR well.
- In implementing a student model, both CBR and jCOLIBRI has its own advantages and drawbacks.
- The system can support formative evaluation in the course Microprocessor Systems by showing patterns of previous cases to the student as their feedback. However the system can only show the cases individually, not as a whole class data.
- This system can help student and lecturer in predicting their final grade, thus an improvement effort can be made accordingly.

Overall, some recommendations in implementing a system for student model in the future are listed below.

1. Teachers/lecturer must define clearly the structure of the lesson in advance. It must be clearly stated when they are going to take score (how many quiz and exams). The class must also have a complete documentation.
2. Complete data of student is needed prior to conducting the class.
3. The existing e-learning system can also be integrated here.
4. As jCOLIBRI has several drawbacks, the system can also be supported with other features outside the framework that are compatible.